\documentclass[10pt,twocolumn,letterpaper]{article}

\usepackage[pagenumbers]{iccv} 

\usepackage{booktabs}  
\usepackage{makecell}  

\usepackage{amsmath}
\usepackage{amssymb}
\usepackage{nicefrac}
\usepackage{placeins}
\newcommand{\cmark}{\textcolor{LimeGreen}{\Large$\checkmark$}}
\newcommand{\xmark}{\textcolor{RedOrange}{\Large$\times$}}

\definecolor{iccvblue}{rgb}{0.21,0.49,0.74}
\usepackage[pagebackref,breaklinks,colorlinks,allcolors=iccvblue]{hyperref}
\usepackage{soul}
\usepackage{graphicx}
\usepackage{amssymb}
\usepackage{algorithm}
\usepackage[noend]{algorithmic}
\usepackage{epigraph}
\usepackage{nicefrac}

\def\paperID{8974} 
\def\confName{ICCV}
\def\confYear{2025}

\title{Occam's LGS: An efficient approach for Language Gaussian Splatting}

\author{Jiahuan Cheng$^{1,2}$ \quad Jan-Nico Zaech$^{2}$ \quad Luc Van Gool$^{2}$ \quad Danda Pani Paudel$^{2}$ \vspace{0.3em} \\
{\normalsize $^1$\begin{tabular}[t]{@{}c@{}}Johns Hopkins University\\Baltimore, United States\end{tabular}} \qquad
{\normalsize $^2$\begin{tabular}[t]{@{}c@{}}INSAIT, Sofia University\\Sofia, Bulgaria\end{tabular}}
}

\begin{document}
\maketitle
\begin{abstract}


Gaussian Splatting is a widely adopted approach for 3D scene representation, offering efficient, high-quality reconstruction and rendering. A key reason for its success is the simplicity of representing scenes with sets of Gaussians, making it interpretable and adaptable. To enhance understanding beyond visual representation, recent approaches extend Gaussian Splatting with semantic vision-language features, enabling open-set tasks. Typically, these language features are aggregated from multiple 2D views, however, existing methods rely on cumbersome techniques, resulting in high computational costs and longer training times.

In this work, we show that the complicated pipelines for language 3D Gaussian Splatting are simply unnecessary. Instead, we follow a probabilistic formulation of Language Gaussian Splatting and apply Occam's razor to the task at hand, leading to a highly efficient weighted multi-view feature aggregation technique. Doing so offers us state-of-the-art results with a speed-up of two orders of magnitude without any compression, allowing for easy scene manipulation. Project Page: \url{https://insait-institute.github.io/OccamLGS/}

\end{abstract}    
\FloatBarrier
\begin{figure}[t] 
    \centering
    \includegraphics[width=\columnwidth, trim={0 380 541 0}, clip]{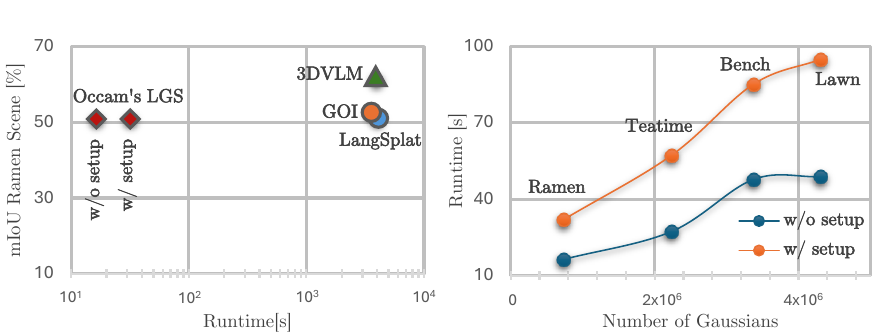} 
    \caption{Occam's LGS performs training-free language feature aggregation in 3D by accurately modeling rendering of Gaussian Splatting, avoiding expensive training in feature space. This improves runtime by two orders of magnitude, while achieving SOTA performance for open-set vision-language tasks and allowing for downstream applications such as scene editing.}
    \label{fig:teaser}
\end{figure}

\section{Introduction}
\label{sec:intro}

3D semantic understanding is a fundamental capability in computer vision that enables machines to interpret and reason about the 3D world, with practical use in robotics~\cite{rosinol20203ddynamicscenegraphs}, autonomous navigation~\cite{zhou2020cylinder3d, tian2024drivevlmconvergenceautonomousdriving}, augmented reality, and interactive systems. While traditional 3D scene understanding~\cite{gupta2015indoor} focuses on extracting geometric structure and predefined information from visual data, newer approaches integrate natural language. This emerging field, known as 3D language understanding~\cite{peng2023openscene3dsceneunderstanding, kerr2023lerf, gu2023conceptgraphsopenvocabulary3dscene, liu2023weakly, ha2022semantic, zhang2023clip, takmaz2023openmask3d, ding2023pla}, aims to establish rich and meaningful connections between linguistic expressions and 3D visual representations such as point clouds, meshes, and RGB-D images.

Recent breakthroughs in 3D scene representation~\cite{nerf, kerbl3Dgaussians, sun2022direct, yu2021plenoxels}, particularly 3D Gaussian Splatting~\cite{kerbl3Dgaussians, Yu2024MipSplatting}, have revolutionized scene reconstruction and rendering by offering an efficient and high-quality approach to modeling 3D environments. While 3D Gaussian Splatting excels at geometric reconstruction and novel view synthesis, its potential for semantic understanding remains largely unexplored. Recent research has begun addressing this limitation by incorporating language features into the Gaussian representations, enabling open-vocabulary scene understanding through natural language queries. 

Vision-language models like CLIP~\cite{caron2021emerging} and LSeg~\cite{ghiasi2022scaling} have demonstrated the power of joint visual-linguistic understanding by training on vast image-text pairs, while foundation models like Segment Anything (SAM)~\cite{kirillov2023segment} and DINO-V2~\cite{oquab2024dinov2} have shown the potential of large-scale visual learning.
Although these models have changed 2D image understanding completely, extending their capabilities to 3D domains remains challenging~\cite{qin2024langsplat, peng20243d, wu2024opengaussian, shi2024language, ye2025gaussian, bhalgat2024n2f2}: directly learning 3D-language representations lacks paired data and 2D vision-language models typically produce expressive high-dimensional features, which poses an often infeasible computational cost when integrated in complex 3D scenes.

Current approaches address this challenge through scene-specific feature autoencoders~\cite{qin2024langsplat} or quantization to create a compact feature space~\cite{shi2024language}. However, these scene-specific compression approaches fundamentally limit scalability and generalization. While effective for individual environments, they create barriers for multi-scene applications by requiring separate training for each scene, increasing both computational and storage overhead. Moreover, these scene-specific solutions make it difficult to transfer knowledge between environments and limit the potential for open-world understanding.

We take a fundamentally different approach by proposing a training-free method, that avoids repeated iteration over views, as well as loss computation in the high-dimensional feature space. Overall, our contribution can be summarized as follows:
\begin{itemize}
    \item A training-free global optimization approach for Language 3D Gaussian Spatting, improving speed by two orders of magnitude.
    \item A reasoning by rendering approach to avoiding expensive feature rendering by following a probabilistic formulation of the Gaussian Splatting forward process.
    \item Compatibility with uncompressed and arbitrary dimensional language features.
\end{itemize}

\begin{table}[h]
    \centering
    \footnotesize
    \begin{tabular}{@{} lcccc|cc}
        \toprule
        & \multicolumn{4}{c}{Requires} & & \\ 
        Method
        & \rotatebox{90}{\parbox{1.8cm}{\raggedright Feature Compression}}
        & \rotatebox{90}{\parbox{1.8cm}{\raggedright Spatial Compression}}
        & \rotatebox{90}{\parbox{1.8cm}{\raggedright Back Propagation}}
        & \rotatebox{90}{\parbox{1.8cm}{\raggedright Scene Specific Model}}
        & \rotatebox{90}{\parbox{1.8cm}{\raggedright Probabilistically Grounded}}        
        & \rotatebox{90}{\parbox{1.8cm}{\raggedright Fast}} \\
        \midrule
        LERF~\cite{kerr2023lerf} & \xmark & \xmark & \cmark & \cmark & \xmark & \xmark \\
        GS-Grouping~\cite{ye2025gaussian} & \cmark & \xmark & \cmark & \cmark & \xmark & \xmark \\
        Feature-3DGS~\cite{zhou2024feature} & \cmark & \xmark & \cmark & \cmark & \xmark & \xmark \\
        LEGaussians~\cite{shi2024language} & \cmark & \xmark & \cmark & \cmark & \xmark & \xmark \\
        GOI~\cite{qu2024goi3dgaussiansoptimizable} & \cmark & \xmark & \cmark & \cmark & \xmark & \xmark \\
        LangSplat~\cite{qin2024langsplat} & \cmark & \xmark & \cmark & \cmark & \xmark & \xmark \\
        FMGS~\cite{zuo2024fmgs} & \xmark & \cmark & \cmark & \cmark & \xmark & \xmark \\
        Occam's LGS & \xmark & \xmark & \xmark & \xmark & \cmark & \cmark \\
        \bottomrule
    \end{tabular}
    \caption{Our approach challenges the necessity of complex modules to lift 2D language features to Gaussian Splats, resulting in a simple and fast, yet probabilistically grounded method.}
    \vspace{-5mm}
    \label{tab:method-comparison}
\end{table}
\section{Related works}
\label{sec:related_works}

\subsection*{3D Open Vocabulary Scene Understanding} Recent works~\cite{zhang2023clip, chen2023clip2scene, peng2023openscene3dsceneunderstanding, tschernezki22neural, goel2023interactive} have explored 2D vision-language models such as CLIP~\cite{Radford2021LearningTV}, OpenSeg~\cite{ghiasi2022scaling}, and OVSeg~\cite{liang2023open}, and foundation models such as SAM~\cite{kirillov2023segment} to ground 2D language-aligned features in 3D space. Initial approaches~\cite{kerr2023lerf, kobayashi2022decomposing, Haque_2023_ICCV} demonstrated the feasibility of distilling language features into NeRF-based representations, though at the cost of slow training and rendering. More recent Gaussian-based methods have improved efficiency while following a common framework: extracting language-aligned feature maps from 2D views and grounding them to 3D space through multi-view optimization. LangSplat~\cite{qin2024langsplat} employs SAM to obtain masks at different granularity, then extracts CLIP image features for these regions. LEGaussians~\cite{shi2024language} combines DINO~\cite{caron2021emerging} and CLIP features to obtain hybrid dense language feature maps and FMGS~\cite{zuo2024fmgs} uses a similar combination to improve robustness. GOI~\cite{qu2024goi3dgaussiansoptimizable} leverages APE~\cite{shen2024aligning} to extract pixel-aligned features with fine boundaries.

However, incorporating the high-dimensional language features into 3D space presents unique challenges. It not only increases memory requirements but also significantly extends training time. Different approaches have been proposed to address this challenge: LangSplat~\cite{qin2024langsplat} employs a scene-specific autoencoder for feature compression, LEGaussians~\cite{shi2024language} uses quantization of semantic features, and Feature-3DGS~\cite{zhou2024feature} employs feature compression with a lightweight decoder. FMGS~\cite{zuo2024fmgs} introduces multi-resolution encoding for efficient semantic embedding and GOI~\cite{qu2024goi3dgaussiansoptimizable} utilizes a trainable feature clustering codebook.

In contrast to these methods, our approach supports arbitrary-dimensional uncompressed features without compromising training speed. Our method preserves semantic meaning without requiring additional preprocessing steps or encoder-decoder structures. This ensures open-vocabulary generalization to the same extent as the underlying feature extractor and enables a range of downstream applications using dynamic scenes or scene editing.

\subsection*{Uplifting 2D Features to 3D} 
While current methods~\cite{qin2024langsplat,shi2024language,peng20243d, kobayashi2022decomposing} focus on optimization-based feature lifting, several approaches also explore direct back-projection of 2D features to 3D space using pixel rays and depth information.
OpenGaussian~\cite{wu2024opengaussian} proposes an instance-level 3D-2D feature association method that links 3D points to 2D masks. Semantic Gaussians~\cite{guo2024semantic} utilizes depth to find corresponding 3D Gaussians along each pixel's ray. PLGS~\cite{wang2024plgsrobustpanopticlifting} initializes semantic anchor points by back-projecting pixels into 3D space using camera parameters. While not focusing on semantic understanding, FlashSplat~\cite{flashsplat} demonstrates that 2D mask rendering is essentially a linear function of Gaussian labels, enabling closed-form optimization through linear programming. 

In contrast, our method accurately models Gaussian Splatting forward rendering to lift 2D language features to 3D, while avoiding back-projection limitations such as occlusion and incomplete depth information. Table~\ref{tab:method-comparison} summarizes the key differences between our method and existing approaches. Unlike previous approaches, our method is training-free and does not require costly semantic feature rendering. While maintaining comparable performance with existing methods, it efficiently completes the lifting of 512-dimensional features to 3D in less than one minute.
\section{Method}
\label{sec:prelim}
\begin{figure*}[t]
    \centering
    \includegraphics[width=0.8\textwidth, trim={20 0 20 0}, clip]{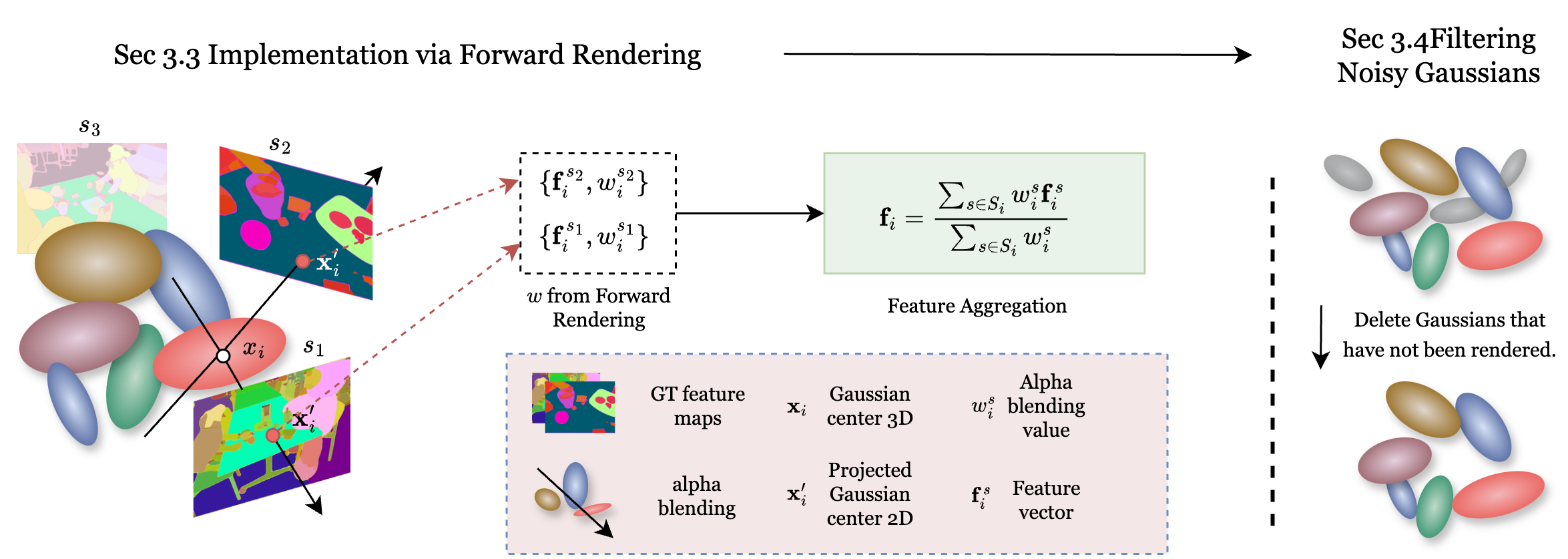}
    \caption{Overview of our method: Occam's LGS consists of two main stages: (1) Forward rendering process of 3D Gaussian Splatting to obtain alpha blending weights $w$, projected positions $x_i'$ for each Gaussian and their corresponding pixels $p_i$ in 2D views, followed by weighted aggregation of multi-view semantic features (Sec. 3.3); and (2) Filtering of noisy Gaussians that remain invisible throughout the rendering process (Sec. 3.4).}
    \label{fig:overview}
\end{figure*}

The following section provides an overview of our method, which establishes a theoretically grounded and efficient approach for uplifting 2D features to 3D space. It introduces the background on Gaussian splatting rendering, our formulation of maximum likelihood features uplifting, and its implementation via forward rendering and postprocessing by filtering. An overview of the flow of our method is visually presented in Figure~\ref{fig:overview} and further detailed in Algorithm~\ref{alg:algorithm1}.

\subsection{Gaussian Splatting Rendering}
3D Gaussian Splatting represents a scene as a collection of anisotropic Gaussians, where each Gaussian is defined by its position \(\mathbf{x} \in \mathbb{R}^3\), covariance \(\Sigma \in \mathbb{R}^6\), color represented by either a 3D color representation \(\mathbf{c} \in \mathbb{R}^3\), or a viewpoint dependent representation as spherical harmonics~(SH) parameters, and opacity \(o \in \mathbb{R}\)
\begin{equation}
    G(\mathbf x) = e^{-\frac{1}{2}(\mathbf{x})^{\intercal}\Sigma^{-1}(\mathbf{x})}.
\end{equation}
The primitive representation of a Gaussian is defined as $G = \{\mathbf{x}, \Sigma, \mathbf{c}, o\}$, and let $\mathcal{G}=\{G_i\}_{i=1}^N$ be the set of Gaussians in a scene. For rendering, we first project each 3D Gaussian onto the image plane, resulting in a 2D Gaussian $G' = \{\mathbf{x'}, \Sigma', \mathbf{c}, o\}$, with $\mathbf{x'} \in \mathbb{R}^2$ as the projected 2D coordinates and $\mathbf{\Sigma'} \in \mathbb{R}^{2\times2}$ as the projected covariance matrix.

To render the color at pixel $\mathbf{p}$, we blend the contributions of all Gaussians along the ray originating from the camera center, say $\mathcal{G}^r = \{G_k\}_{k=1}^R$, where $R$ is the total number of Gaussians along the ray. To achieve efficient rendering in the Gaussian Splatting framework, this rendering process is implemented through a tile-based sorting strategy, which utilizes the approximation of Gaussians by a spatially truncated distribution. For this purpose, the image is partitioned into tiles, typically spanning 16x16 pixels. Each 2D Gaussian is assigned to all tiles intersecting with a bounding box placed at the 3$\sigma$ boundary. Within each tile, Gaussians are sorted by depth from the camera center. These sorted Gaussians represent the collection of Gaussians that potentially contribute to any pixel within that tile, which are then rendered sequentially, starting closest to the camera. The tiling approach implements an efficient ray-Gaussian intersection test: if a Gaussian's bounding box intersects a tile, it is a candidate to contribute to the rays passing through that tile's pixels.

For the \(i\)-th Gaussian in the sorted list, its contribution to a pixel $\mathbf{p}$ is defined as $\alpha_i = o_i G_i(\mathbf{x}_i')$, representing its influence. When projecting all relevant Gaussians, the color of the pixel $\mathbf{p}$ is computed through a weighted blending process given by
\begin{equation}
 \mathcal{C}(\mathbf{p}) = \psi_c (\mathbf{p}, \mathcal{G}) = \sum_{i=1}^{R} \alpha_i \mathbf{c}_i \prod_{j=1}^{i-1} (1 - \alpha_j),
 \label{eq:colorR}
\end{equation}
where, $\psi_c(,.,)$ is the color rendering function. Specifically, the contribution of $G_i$ to pixel $\mathbf{p}$ is computed as
\begin{equation}
    \alpha_i = \mathbf{o}_i \cdot e^{-\frac{1}{2}((\mathbf{p} - \mathbf x_i ')^{\intercal}\Sigma^{-1}(\mathbf{p} - \mathbf x_i ')}.
    \label{eq:featureR}
\end{equation}

With the goal of lifting multi-view 2D features into 3D space, we extend each Gaussian representation to include a high-dimensional semantic feature vector $\mathbf{f} \in \mathbb{R}^{d}$. Thus, each augmented 3D Gaussian $G_i$ learns its own semantic feature $\mathbf{f_i}$. Similar to color blending, the semantic feature at pixel $\mathbf{p}$ is computed through alpha-weighted blending
\begin{align}
\mathcal{F}(\mathbf{p}) = \psi_f (\mathbf{p}, \mathcal{G}) & = \sum_{i=1}^{R} \alpha_i \mathbf{f}_i\prod_{j=1}^{i-1} (1 - \alpha_j) \\
&= \sum_{i=1}^{R} w_i \mathbf{f}_i  
\label{eq:featureR}
\end{align}
where $\psi_f$ is the feature rendering function and $w_i = \alpha_i\prod_{j=1}^{i-1} (1 - \alpha_j)$ represents the alpha blending weight for Gaussian $i$. 

It is important to note that explicit rendering of semantics according to Equation~\eqref{eq:featureR} is expensive both in memory and computation when compared to the rendering of color in Equation~\eqref{eq:colorR}. This is due to the considerably higher dimensionality of the features $d>>3$, which prevents the GPU from fitting all required data in the shader L1 cache~\cite{qin2024langsplat}.

\subsection{Maximum Likelihood Feature Uplifting}
We deeply ground our approach to feature uplifting in the principles of 3DGS and probabilistic modeling. While conventional semantic feature learning methods~\cite{wu2024opengaussian, guo2024semantic} rely on back-projection, where 2D pixel-space language features are projected back to 3D space through ray-tracing, we leverage the underlying mathematical structure of Gaussian Splatting's forward rendering process to establish a more precise and theoretically sound feature lifting mechanism and relate it to forward modeling.

We formulate the feature uplifting problem as a Maximum-a-Posteriori (MAP) estimation problem. Given 2D feature maps $\mathbf{F}^{2D}$, the MAP estimator of 3D features $\mathbf{f}$ is:
\begin{equation}
\arg\max_\mathbf{f} p(\mathbf{f}|\mathbf{F}^{2D}) = \arg\max_\mathbf{f} p(\mathbf{F}^{2D}|\mathbf{f})\frac{p(\mathbf{f})}{p(\mathbf{F}^{2D})}
\end{equation}
For language gaussian splatting with uniformly distributed language features $p(\mathbf{f})$, which we assume for the efficient feature space of CLIP, the estimator simplifies to the Maximum Likelihood (ML) formulation $\arg\max_\mathbf{f} p(\mathbf{F}^{2D}|\mathbf{f})$.

Under a Gaussian noise model $\epsilon_{p,v}$ for language features and conditional independence of pixels, the rendered 2D feature at pixel $p$ in view $v$ is $\mathbf{f}_{p,v}^{2D} = \sum_{k=1}^R w_{k,p,v}\mathbf{f}_k + \epsilon_{p,v}$. This leads to the weighted aggregation formula for the 3D feature of each Gaussian
\begin{equation}
\boxed{
\mathbf{f}_i = \frac{\sum\limits_{s \in S_i} w_i^s \mathbf{f}_i^s}{\sum\limits_{s \in S_i} w_i^s}
}
\label{eq:wtd.agg}
\end{equation}
where $S_i$ is the set of views where Gaussian $i$ is visible, $w_i^s$ represents the contribution (alpha blending weight) of Gaussian $i$ at its center projection in view $s$, and $\mathbf{f}_i^s$ is the 2D feature at Gaussian $i$'s center projected pixel in view s.
This formula emerges naturally from our probabilistic framework and provides a solution that weights each view's contribution according to the Gaussian's influence in that view. Views, where a Gaussian is more prominent (with a higher alpha blending weight), contribute more significantly to its 3D feature, addressing the challenge of noisy or partially occluded perspectives. We provide complete step-by-step derivation in the supplementary material.

\subsection{Implementation via Forward Rendering}
Our weighted aggregation formula directly maps to the forward rendering process, enabling an efficient implementation that avoids complex back-projection or iterative optimization.
Given a trained visual Gaussian model, for each Gaussian $G_i$ we perform the following steps:
\begin{enumerate}
    \item Identify the set of views $S_i$ where the Gaussian $G_i$ is visible (within the view frustum).
    \item Determine the Gaussian's center projected pixel location $\mathbf{p}_i^s = \lfloor\mathbf{x}_i'\rfloor$ in each view $s$.
    \item Record the alpha blending weight $w_i^s$ and extract the 2D feature $\mathbf{f}_i^s$ at each center projected pixel.
    \item Compute the weighted average following Equation~\ref{eq:wtd.agg}.
\end{enumerate}
This implementation leverages the existing tile-based rendering architecture, allowing us to record projection locations and alpha values during the optimized standard rendering pass. The forward feature collection approach offers several key advantages; First, the feature uplifting requires only a single forward pass through the views, making it computationally efficient. Second, view frustum culling and tile-based sorting naturally handle occluded Gaussians, ensuring we only process visible features. Finally, by explicitly recording projection locations during rendering, we establish direct and precise correspondence between 2D features and 3D Gaussians, eliminating the need for costly back-projection or iterative optimization methods.

\subsection{Filtering Noisy Gaussians}
After the forward rendering pass, we identify and filter out Gaussians that have no contributions across all views. This filtering effectively reduces noise while reducing the number of Gaussians significantly. Besides, this statistical outlier elimination is consistent with our probabilistic framework, as Gaussians with minimal visibility have high uncertainty in their feature estimates. Through this process, we obtain a refined set of semantic Gaussians $\mathcal{G}^s = \{{G_i^s}\}_{i=1}^M$, where $M \ll N$, representing only the Gaussians that meaningfully contribute to the scene's semantic representation.

\begin{algorithm}
\caption{Occam's LGS}
\label{alg:algorithm1}
\small
\begin{algorithmic}[1]
\REQUIRE
\STATE Training camera views $\mathcal{S} = \{s_1, ..., s_n\}$
\STATE Set of trained 3D Gaussians $\mathcal{G} = \{G_1, ..., G_m\}$
\STATE 2D language feature maps $\mathcal{F} = \{\mathbf{f}^{s_1}, ..., \mathbf{f}^{s_n}\}$
\ENSURE

\STATE \textbf{Initialize:}
\FOR{each Gaussian $G_i \in \mathcal{G}$}
    \STATE $\mathbf{f}_i \leftarrow 0$ \COMMENT{Accumulated features}
    \STATE $w_i \leftarrow 0$ \COMMENT{Accumulated weights}
\ENDFOR

\STATE \textbf{Feature Uplifting Phase:}
\FOR{each camera view $s \in \mathcal{S}$}
    \FOR{each Gaussian $G_i$ in view frustum}
        \STATE $w_i^s, \mathbf{p}_i^s \leftarrow \text{Render}(G_i|s)$ \COMMENT{Project and Render}
        \STATE $\mathbf{f}_i^s \leftarrow$ 2D feature map for view $s$
        \STATE $\mathbf{f}_i \leftarrow \mathbf{f}_i + w_i^s \mathbf{f}_i^s$ \COMMENT{Sampled feature}
        \STATE $w_i \leftarrow  w_i + w_i^s$ \COMMENT{Alpha blending value}
    \ENDFOR
\ENDFOR
\FOR{each Gaussian $G_i \in \mathcal{G}$}
    \STATE $\mathbf{f}_i \leftarrow \mathbf{f}_i/w_i$ \COMMENT{Feature Normalization}
\ENDFOR

\RETURN $\mathcal{G}$
\end{algorithmic}
\end{algorithm}
\section{Experiments}
\label{sec:experiment}
We evaluate our method against other state-of-the-art 3D Language Gaussian Splatting methods on open vocabulary segmentation and localization tasks, as well as compare their computational efficiency metrics. Additionally, we analyze how our method scales across different feature dimensions, examining the trade-offs between dimensionality, computational requirements, and segmentation quality.

\begin{figure}[t]
    \centering
    \includegraphics[width=0.88\textwidth]{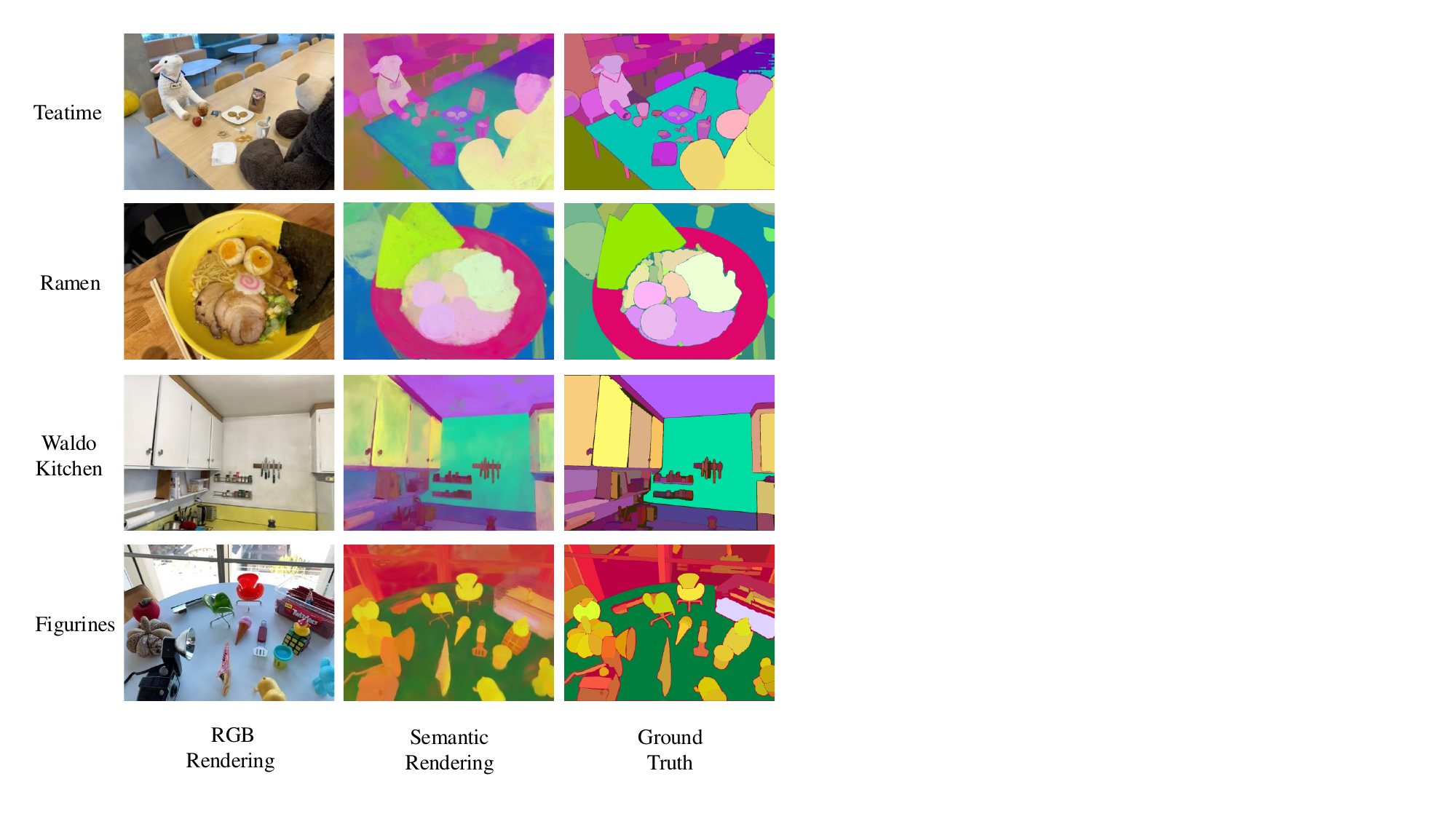}
    \caption{PCA visualization of semantic features on LERF Dataset. (a) Original RGB rendering. (b) PCA visualization of our rendered semantic feature maps. (c) Initial semantic feature maps extracted using CLIP from SAM segmentations. }
    \label{fig:lerf-vis}
\end{figure}

\subsection{Dataset} To assess the effectiveness of our method, we evaluate it on two widely used datasets: LERF~\cite{kerr2023lerf} and 3D-OVS~\cite{liu2023weakly}. The LERF dataset features everyday indoor scenes with diverse objects and we perform quantitative evaluations on the commonly used scenes Ramen, Figurines, Teatime, and Waldo Kitchen according to the LERF evaluation protocol~\cite{kerr2023lerf}. For these scenes, we evaluate both open-vocabulary segmentation (mIoU) and localization accuracy. The 3D-OVS dataset contains long-tail objects in diverse backgrounds, and we evaluate segmentation performance on five scenes: sofa, lawn, bed, room, and bench. For evaluation, we utilize the annotations provided by~\cite{qin2024langsplat} for both datasets.

\subsection{Implementation Details}
Our implementation of Occam's LGS is built on gsplat~\cite{ye2024gsplatopensourcelibrarygaussian}, a CUDA-accelerated Gaussian rendering library that supports high-dimensional feature rendering. For each scene, we train a 3D Gaussian Splatt for 30,000 iterations using the common Gaussian Splatting~\cite{kerbl3Dgaussians} parameters, resulting in an average number of 2,500,000 Gaussians for RGB representation across scenes. To ensure a fair comparison, we follow the 2D language feature map generation method from~\cite{qin2024langsplat}. For language feature extraction and segmentation, our pipeline integrates OpenCLIP ViT-B/16~\cite{Radford2021LearningTV} and SAM ViT-H~\cite{kirillov2023segment}. To capture semantic information hierarchically, we generate 2D feature maps at three levels:  whole, part, and subpart, where each level represents distinct semantic granularity. Following the feature extraction, we use our method to uplift the 2D language features to 3D scenes using training views only, and evaluate the results on separate testing views, withheld during training of the semantic representation. All experiments and runtime measurements are conducted on a single NVIDIA A6000 GPU. 

For the LERF dataset, we follow the LERF~\cite{kerr2023lerf} evaluation protocol, computing relevancy scores between query text and rendered 2D feature maps with a mask threshold of 0.5. For 3D-OVS, we first normalize the rendered feature maps to ensure comparable relevancy scores across classes, then apply dynamic thresholding detailed in the supplementary material. During thresholding, we reject spurious masks with areas smaller than 0.005\% or larger than 0.90\% of the image and set an object stability threshold of 0.4.

\subsection{Open Vocabulary Semantic Segmentation}
\begin{figure}[h]
    \centering
    \includegraphics[trim={3cm 1cm 1cm 1cm}, clip, width=0.52\textwidth]{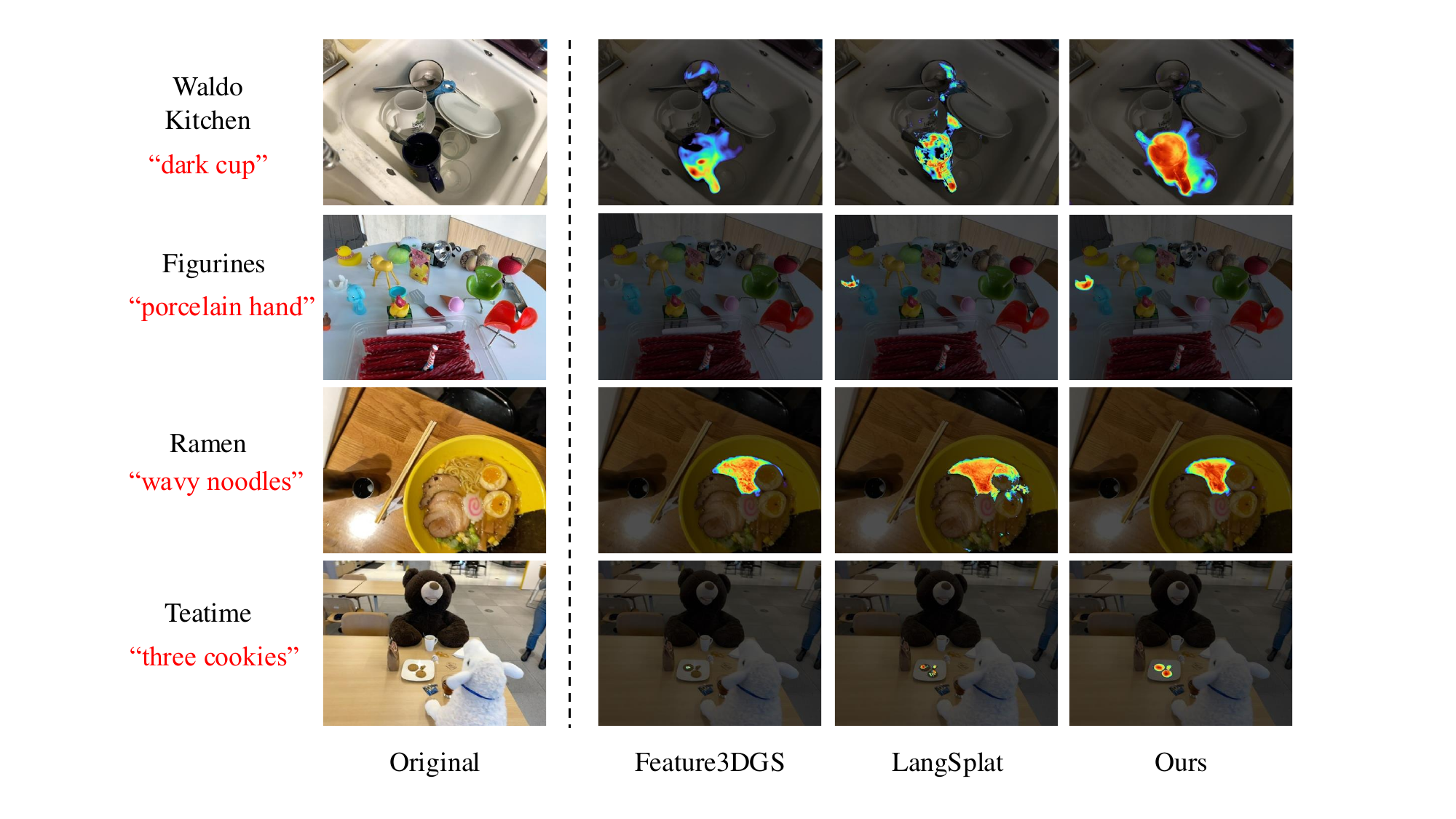}
    \caption{Qualitative comparison of relevancy score visualization. }
    \label{fig:comparison}
\end{figure}

Given a 3D scene and a language query describing an object or region of interest, the task of open vocabulary semantic segmentation is to identify and generate masks for all relevant regions that match a query description. We evaluate our approach against existing methods on two datasets: LERF and 3D-OVS, with quantitative results shown in Tables \ref{tab:lerf-mIoU} and \ref{tab:3dovs-mIoU} respectively and qualitative results shown in Figure \ref{fig:lerf-vis} and \ref{fig:comparison}. On the LERF dataset, under identical experimental settings as LangSplat, our approach demonstrates a substantial improvement of 9.9\% mIoU. Notably, our method shows advantages, especially in complex scenes with sparse view coverage. An example of this is Waldo Kitchen, where the camera captures an entire room rather than focusing on a specific area that is fully surrounded by camera views. 

On 3D-OVS, our method achieves state-of-the-art results, with an improvement of $1.6\%$ mIoU over the current SOTA LangSplat~\cite{qin2024langsplat}. Performance gains of 3D-OVS are less pronounced compared to LERF~\cite{kerr2023lerf}, likely since this dataset contains simpler scenes, focused on a single region with good view coverage and fewer objects, where existing methods have already saturated in performance.

\begin{table}[t]
    \centering
    \footnotesize
    \begin{tabular}{lccccc}
        \toprule
        \textbf{Method}& 
        \hspace{-2pt}\textbf{Ramen}\hspace{-2pt} & \hspace{-2pt}\textbf{Figurines}\hspace{-2pt} & \hspace{-2pt}\textbf{Teatime}\hspace{-2pt} & \hspace{-2pt}\makecell{\textbf{Waldo}\\\textbf{Kitchen}}\hspace{-2pt} & \hspace{-2pt}\textbf{Overall}\hspace{-2pt} \\
        \midrule
        LSeg & 7.0 & 7.6 & 21.7 & 29.9 & 16.6 \\
        LERF & 28.2 & 38.6 & 45.0 & 37.9 & 37.4 \\
        GS-Grouping & 45.5 & \underline{60.9} & 40.0 & 38.7 & 46.3 \\
        Feature-3DGS & 43.7 & 58.8 & 40.5 & 39.6 & 45.7 \\
        LEGaussians & 46.0 & 60.3 & 40.8 & 39.4 & 46.9 \\
        GOI & $\mathbf{52.6}$ & $\mathbf{63.7}$ & 44.5 & 41.4 & 50.6 \\
        LangSplat & $\underline{51.2}$ & 44.7 & $\underline{65.1}$ & $\underline{44.5}$ & $\underline{51.4}$ \\
        Ours & 51.0 & 58.6 & $\mathbf{70.2}$ & $\mathbf{65.3}$ & $\mathbf{61.3}$ \\
        \bottomrule
    \end{tabular}
    \caption{Comparison of average IoU on LERF Dataset.}
    \label{tab:lerf-mIoU}
\end{table}
\begin{table}[t]
    \centering
    \footnotesize
    \begin{tabular}{lcccccc}
        \toprule
        \hspace{-2pt}\textbf{Method}\hspace{-2pt} &
        \hspace{-2pt}\textbf{Bed}\hspace{-2pt} &
        \hspace{-2pt}\textbf{Bench}\hspace{-2pt} &
        \hspace{-2pt}\textbf{Room}\hspace{-2pt} &
        \hspace{-2pt}\textbf{Sofa}\hspace{-2pt} &
        \hspace{-2pt}\textbf{Lawn}\hspace{-2pt} &
        \hspace{-2pt}\textbf{Overall}\hspace{-2pt}\\
        \midrule
        \hspace{-2pt}LSeg & 56.0 & 6.0 & 19.2 & 4.5 & 17.5 & 20.6\\
        \hspace{-2pt}LERF & 73.5 & 53.2 & 46.6 & 27 & 73.7 & 54.8\\
        \hspace{-2pt}GS-Grouping & 83.0 & 91.5 & 85.9 & 87.3 & 90.6 & 87.7\\
        \hspace{-2pt}Feature-3DGS\hspace{-3pt} & 83.5 & 90.7 & 84.7 & 86.9 & 93.4 & 87.8\\
        \hspace{-2pt}LEGaussians & 84.9 & 91.1 & 86.0 & 87.8 & 92.5 & 88.5\\
        \hspace{-2pt}GOI & 89.4 & 92.8 & 91.3 & 85.6 & 94.1 & 90.6\\
        \hspace{-2pt}LangSplat & $\underline{92.5}$ & $\underline{94.2}$ & $\underline{94.1}*$ & $\mathbf{90.0}$ & $\underline{96.1}$ & $\underline{93.4}$\\
        Ours & $\mathbf{96.8}$ & $\mathbf{95.8}$ & $\mathbf{96.5}*$ & $\underline{88.8}$ & $\mathbf{97.0}$ & $\mathbf{95.0}$\\
        \bottomrule
    \end{tabular}
    \caption{Comparison of average IoU on 3D-OVS Datasets. The utilized image resolution is 1440 $\times$ 1080. Methods with * are evaluated on testing-data that has an annotation error in the 3D-OVS testset corrected.}
    \label{tab:3dovs-mIoU}
\end{table}

\subsection{Localization}
We further evaluate on the language-guided localization task, where, given a language query, the pixel with the highest relevancy score is selected, following~\cite{kerr2023lerf}. Benchmarking against existing methods, our approach achieves competitive localization accuracy, as shown by Table \ref{tab:lerf-loc-acc}.

Adding feature compression as in existing methods, increases localization accuracy of our approach to 84.1. This is due to the metric only measuring the closest matching pixel, while ignoring the feature distribution and object shape, making it prone to noise. We provide detailed experiments in this setting in the supplementary material.

\begin{table}[t]
    \centering
    \footnotesize
    \begin{tabular}{lccccc}
    \toprule
        \textbf{Method}& 
        \hspace{-2pt}\textbf{Ramen}\hspace{-2pt} & \hspace{-2pt}\textbf{Figurines}\hspace{-2pt} & \hspace{-2pt}\textbf{Teatime}\hspace{-2pt} & \hspace{-2pt}\makecell{\textbf{Waldo}\\\textbf{Kitchen}}\hspace{-2pt} & \hspace{-2pt}\textbf{Overall}\hspace{-2pt} \\
        \midrule
        LSeg & 14.1 & 8.9 & 33.9 & 27.3 & 21.1 \\
        LERF & 62.0 & 75.0 & 84.8 & 72.7 & 73.6 \\
        GS-Grouping & 68.6 & 75.0 & 74.3 & 88.2 & 76.5 \\
        Feature-3DGS & 69.8 & 77.2 & 73.4 & 87.6 & 77.0 \\
        LEGaussians & 67.5 & 75.6 & 75.2 & 90.3 & 77.2 \\
        GOI & 75.5 & $\mathbf{88.6}$ & 82.9 & $\underline{90.4}$ & $\mathbf{84.4}$ \\
        LangSplat & $\underline{73.2}$ & $\underline{80.4}$ & $\underline{88.1}$ & $\mathbf{95.5}$ & $\underline{84.3}$ \\
        Ours & $\mathbf{74.7}$ & $\underline{80.4}$ & $\mathbf{93.2}$ & 81.8 & 82.5 \\
        \bottomrule
        \end{tabular}
        \caption{Comparison of localization accuracy on LERF Datasets.}
        \label{tab:lerf-loc-acc}
\end{table}

\subsection{Robustness to Feature Dimension Variations}
\label{sec:featdim}
In the following, we demonstrate the advantages of different feature dimensions that Occam's LGS enables and assess the respective scalability and performance. For these experiments, we 1) used an autoencoder to compress the CLIP features, 2) uplifted them to 3D using our method, 3) rendered the compressed features, and 4) decompressed features to 512D using the autoencoder. Table~\ref{tab:lerf-featdim-mIoU} shows the IoU across different feature dimensions on LERF datasets. Notably, segmentation performance improves to 64.5~\%mIoU with 64D features, outperforming the LangSplat~\cite{qin2024langsplat} baseline by $13.1\%$~mIoU. This confirms that feature compression filters noise by limiting the feature space to objects present in the 3d scene, at the cost of limited scene editing capability. Since our method supports arbitrary feature dimensions, selecting this operating point is straightforward and can be used depending on the task at hand.

\begin{table}[t]
    \centering
    \footnotesize
    \begin{tabular}{lccccc}
        \toprule
        \textbf{Dimensions}& 
        \hspace{-2pt}\textbf{Ramen}\hspace{-2pt} & \hspace{-2pt}\textbf{Figurines}\hspace{-2pt} & \hspace{-2pt}\textbf{Teatime}\hspace{-2pt} & \hspace{-2pt}\makecell{\textbf{Waldo}\\\textbf{Kitchen}}\hspace{-2pt} & \hspace{-2pt}\textbf{Overall}\hspace{-2pt} \\
        \midrule
        16-D & 48.4 & 57.7 & 75.9 & 59.9 & 60.5 \\
        32-D & \textbf{53.6} & 61.8 & 78.6 & 62.6 & 64.1 \\
        64-D & 51.7 & \textbf{63.4} & \textbf{79.0} & 63.7 & \textbf{64.5} \\
        128-D & 52.1 & 62.4 & 74.3 & 62.6 & 62.9 \\
        256-D & 53.0 & 62.6 & 74.3 & 61.1 & 62.7 \\
        512-D & 51.0 & 58.6 & 70.2 & \textbf{65.3} & 61.3 \\
        \bottomrule
    \end{tabular}
    \caption{Average mIoU with varying feature dimensions on LERF scenes.}
    \label{tab:lerf-featdim-mIoU}
\end{table}

\subsection{Computational Efficiency}
We evaluate the computational efficiency of our method against existing approaches in Figure~\ref{fig:teaser} measuring their semantic Gaussian training times. Similar to ours, LangSplat and GOI both allow the integration of pre-trained 3D Gaussian models, while 3DVLM requires joint uplifting and reconstruction. 
LangSplat, GOI, and 3DVLM take 67, 60, and 65 minutes to uplift three levels of features respectively~\cite{peng20243dvisionlanguagegaussiansplatting}. With our approach, language features can be uplifted in 16.4 seconds without setup, and 31.9 seconds with setup (checkpoint loading and saving).
Overall, our method, achieves feature uplifting two orders of magnitude faster than existing methods, while delivering superior performance across test datasets and avoiding feature compression.

We assess our method's scalability through a thorough analysis of runtime, storage requirements, rendering speed, and GPU memory to uplift one feature level across four scenes of varying complexity. Additionally, we evaluate scaling across different feature dimensions, providing insights into the range of practical deployment scenarios.

\begin{figure}[t]
    \centering
    \includegraphics[width=1\textwidth, trim={0 225 150 0}, clip]{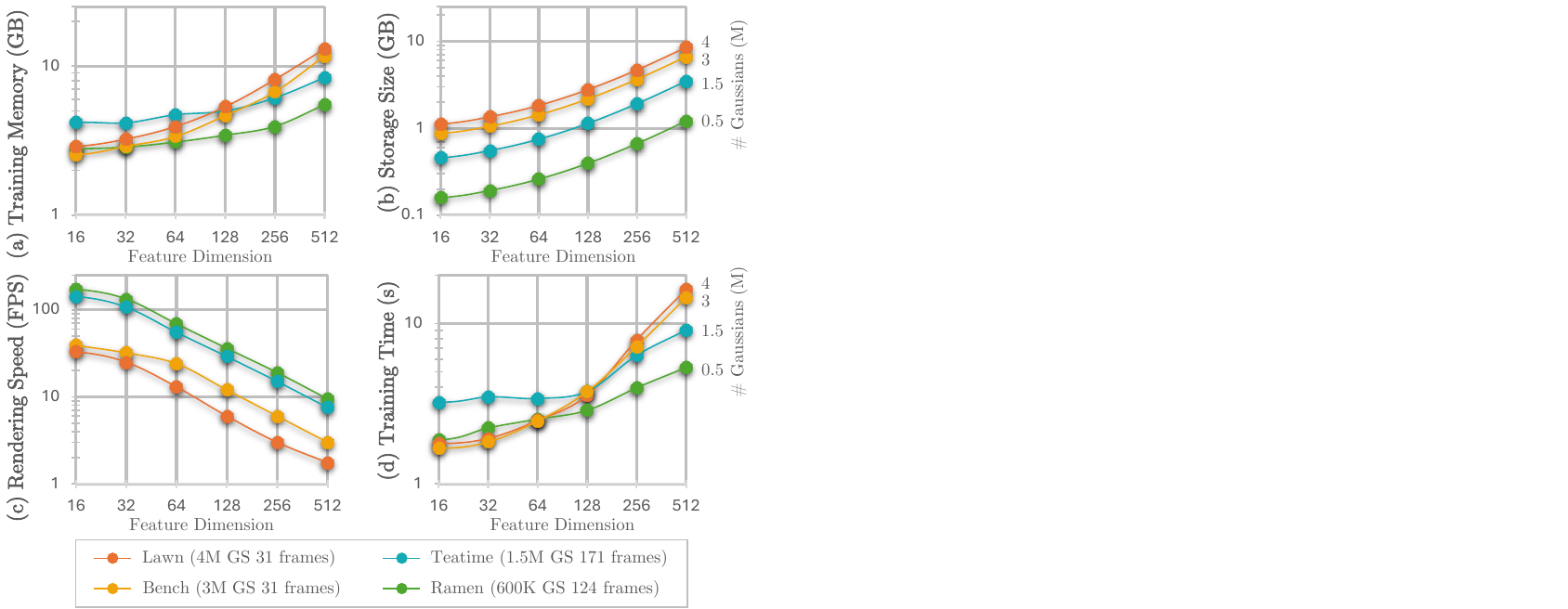}
    \caption{Comprehensive analysis of computational efficiency metrics across different feature dimensions, showing rendering speed (FPS), storage size, GPU memory usage, and runtime performance as Gaussian size and frame numbers vary.}
    \label{fig:comp_efficiency}
\end{figure}

The results depicted in Figure~\ref{fig:comp_efficiency}(d) demonstrate that our method scales well and maintains high computational efficiency even with large scenes containing up to 4 Million Gaussians, indicating its practical scalability across varying scene complexities. The memory requirement, shown by Figure~\ref{fig:comp_efficiency}(a) stays below 14GB for 512D CLIP features, which is met by most modern GPUs.  A key advantage of our method is its flexibility to support features of any dimension, enabling the selection of the optimal feature size for specific applications. From Figure~\ref{fig:comp_efficiency}(c), we observe that rendering speed decreases as feature dimension increases, presenting a clear trade-off between feature expressiveness and computational performance. Storage requirements similarly scale with feature dimensionality, shown by Figure~\ref{fig:comp_efficiency}(b). These findings highlight the adaptability of our approach, enabling users to balance quality, memory usage, and rendering speed based on their available resources and application requirements.


\subsection{Ablation Study}

\begin{figure}[tb]
    \centering
    \includegraphics[trim={3cm 7cm 0cm 1cm}, clip, width=0.55\textwidth]{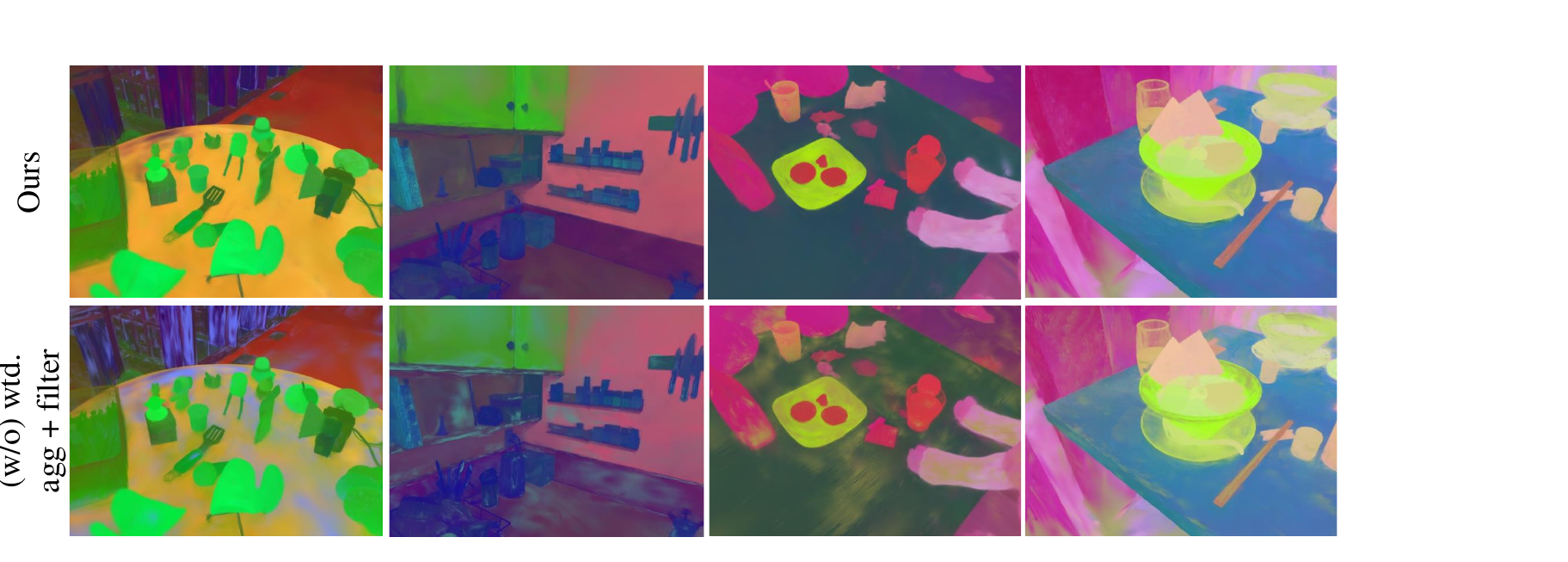}
    \caption{Qualitative results from our ablation study. Top: feature map visualization using our full method with filtering and weighted aggregation. Bottom: baseline results without either component.}
    \label{fig:ablation}
\end{figure}

\begin{table}[t]
   \centering
   \footnotesize
   \begin{tabular}{lc c c c c}
       \toprule
        \textbf{Method} & \hspace{-2pt}\textbf{Ramen}\hspace{-2pt} & \hspace{-2pt}\textbf{Figurines}\hspace{-2pt} & \hspace{-2pt}\textbf{Teatime}\hspace{-2pt} & \hspace{-2pt}\makecell{\textbf{Waldo}\\\textbf{Kitchen}}\hspace{-2pt} & \hspace{-2pt}\textbf{Overall}\hspace{-2pt} \\
        \midrule
        Ours & 51.0 & 58.6 & 70.2 & 65.3 & 61.3 \\
        (w/o) wtd. agg & 45.3 & 56.7 & 70.0 & 64.8 & 59.9  \\
       (w/o) filter& 47.9 & 57.1 & 67.7 & 62.3 & 58.8  \\
        \makecell{(w/o) wtd. agg \\+ filter} & 44.1 & 54.9 & 69.9 & 59.6 & 57.1  \\
       \bottomrule
   \end{tabular}
   \caption{Ablation study on the impact of filtering and weighted aggregation on feature learning. Evaluate on LERF dataset and use average IoU as the metric for comparison.}
   \label{tab:timing_ablation}
\end{table}
We validate the key components of our method with an ablation study depicted in Table \ref{tab:timing_ablation} and Figure \ref{fig:ablation}. First, we ablate our fundamental choice to model the forward rendering process with the weighted feature aggregation strategy. Removing it, while keeping the Gaussian filtering process (w/o wtd. agg), shows that weighted aggregation improves the fusion of multi-view features due to accurate modeling. 

Second, we analyze the impact of filtering redundant Gaussians that are not activated during rendering. Removing filtering (w/o filter) decreases performance, which supports the choice to filter inactive Gaussians. This is because our method uplifts 2D features to a Gaussian only if its center projects onto the corresponding pixel, making it efficient but sparse for occluded Gaussians. Since we threshold occluded Gaussians with negligible influence $w_i^s<10^{-4}$ during training and inference, visual noise can occur. This happens because occluded Gaussians with minimal central influence may still affect neighboring pixels during rendering, but without receiving proper feature updates due to our sparse projection approach. This feature update-rendering influence mismatch creates semantic inconsistencies, so filtering occluded Gaussians improves both consistency and computational efficiency.


Finally, we compare to a baseline not using weighting and filtering(w/o wtd. agg + filter) and thus, perform basic back-projection. This reduces performance by $4.2\%$ and altogether demonstrates that both components contribute meaningfully to our method's performance.

\subsection{Application: Object Insertion}
\begin{figure}[tb]
    \centering
    \includegraphics[trim={4.8cm 6cm 2cm 1cm}, clip, width=0.55\textwidth]{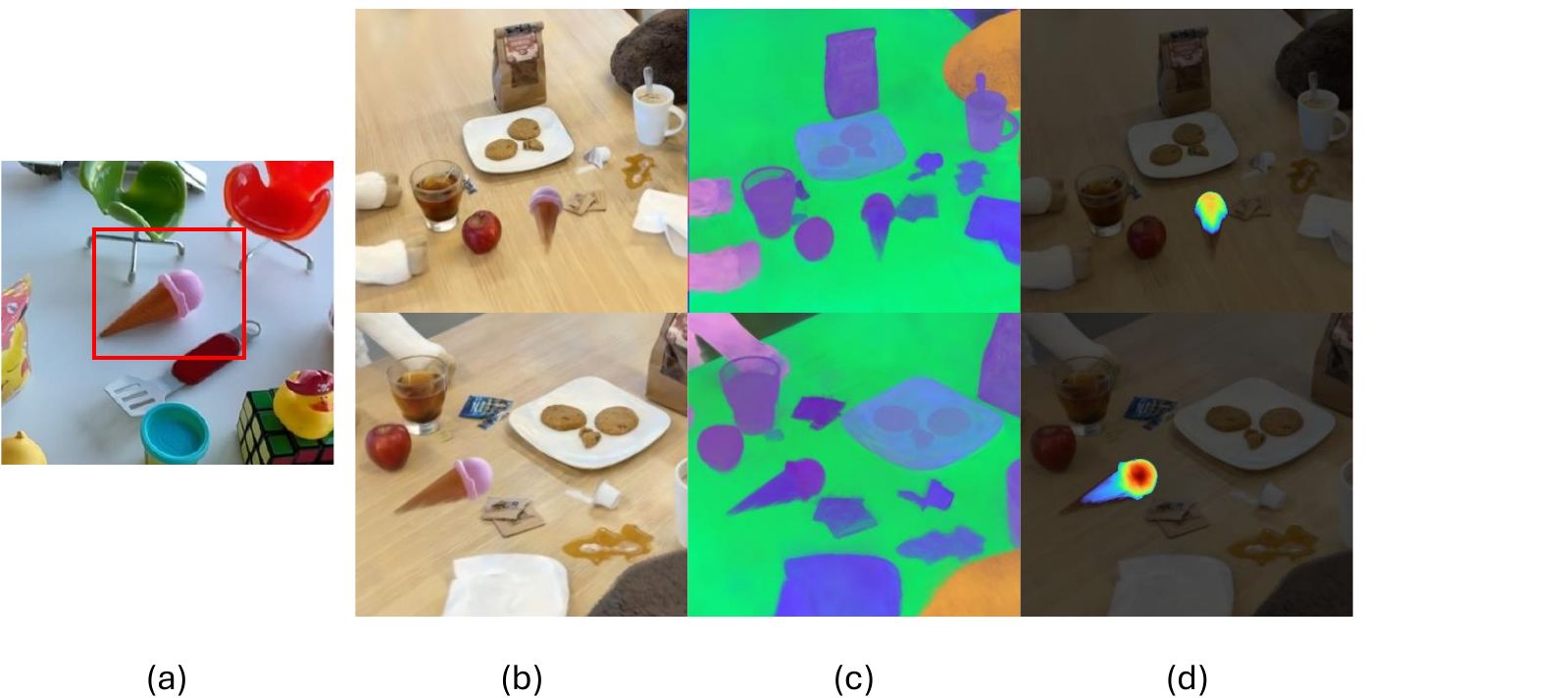}
    \caption{Visualization of Object Insertion: (a) Object Selection: An object is selected in the Figurines scene (b) Object Insertion: The object is transferred and integrated into the Teatime scene (c) Feature Analysis: PCA visualization of the feature map to show the object's integration in the semantic space (d) Query Visualization: Heatmap showing the response to the query ``pink icecream".}
    \label{fig:obj-insert}
\end{figure}

Our method's key advantage is integrating dense 3D language features into Gaussian Splatting with a simple representation. This allows us to profit from their strong open-set generalization capability and the advances in 2D language feature representations. This has the potential to enable a range of downstream 3D tasks, especially related to scene editing and interaction.

We demonstrate this capability on object insertion, where an object is taken from an existing 3D Gaussian Splat and integrated into another scene. While the current state-of-the-art LangSplat would require finetuning of the scene-specific language autoencoder, as well as the retraining of the whole feature embedding, our representation can be augmented with the new object similarly to any other Gaussian splat by simple copy-paste editing. As illustrated in Figure \ref{fig:obj-insert}, we select and extract an ice cream cone (represented by its Gaussians) from the Figurines scene of LERF~\cite{kerr2023lerf}. By simply copying the object's Gaussians together with their parameters and semantic features, the new object seamlessly integrates into the Teatime scene while preserving its semantic features. Then, we query the modified scene with "pink ice-cream". The resulting relevancy heatmap in Figure~\ref{fig:obj-insert})~(d) confirms that our method maintains the semantic understanding of the inserted object in its new context. This demonstrates that our approach enables direct semantic feature manipulation without scene-specific dimensional reduction, offering flexibility for practical editing while preserving semantic querying capabilities.

\section{Discussion and Conclusion}
In this work, we presented a theoretically grounded and optimization-free approach for efficient Language Gaussian Splatting. The proposed method shows state-of-the-art results with a single iteration over all views and avoids rendering and loss computation in the feature space.

Beyond the immediate performance gains, our approach opens up new possibilities for how we approach 3D scene understanding and manipulation. The ability to reason directly in language feature space, without compression, enables practical applications like object insertion and scene registration that were previously computationally prohibitive. Overall, we demonstrated that:
\begin{itemize}
    \item High-dimensional language features can directly be used in Gaussian splatting by utilizing a training-free optimization approach.
    \item Accurate modeling of the rendering process enables the backproject of highly relevant features from 2D to 3D.
    \item Directly projecting high-dimensional features allows us to use scene editing approaches previously limited to 3D Gaussian Splatting.
\end{itemize}
Based on the experiments and results presented in this work, we can answer the following questions about our work and its relation to existing works using feature compression:

\noindent\textbf{Is feature compression or spatial mapping necessary for Language Gaussian Splatting?} No. Avoiding the explicit rendering of language features and loss computation in the 2D space together with a training-free approach makes the direct use of high-dimensional language features feasible. 

\noindent\textbf{Should we stop using low-dimensional Language Gaussian Splatting?} No. Low-dimensional features are justified in certain scenarios despite constraints. Though our method enables the direct use of uncompressed features, low-dimensional alternatives can reduce memory usage, and accelerate 2D feature map rendering, and filter noise. However, it needs to be considered that scene-specific compression approaches may contradict generalization across scenes.

\noindent\textbf{Are there applications where low-dimensional language features are beneficial?} Yes. For scenarios where the domain of scenes is constrained, no scene editing is required, and storage and rendering cost exceeds the cost of training a feature compressor, low dimensional features are advantageous. Thus, the optimal feature resolution is strongly application-dependent and scales, especially in the domain of potential scenarios.

Overall, it is important to acknowledge that our approach is complementary to existing works in Language Gaussian Splatting and enables a new direction to train the 3D language representation. Due to this complementary nature, Occam's LGS can be combined with existing feature compression approaches, with the potential to speed up rendering and reduce storage costs. The right combination of approaches depends on the application and expected use case of the 3D representation, which is a direction we hope future work will explore.
{
    \small
    \bibliographystyle{ieeenat_fullname}
    \bibliography{main}
}

\clearpage
\setcounter{page}{1}
\maketitlesupplementary

\section{3D Feature Uplifting as Maximum Likelihood Estimation}
Our weighted feature uplifting formula is strongly grounded in the forward rendering process from a probabilistic perspective: Given the task of estimating 3d features $\mathbf{f} \in \mathbb{R}^d$ from a set of observed 2D features $\mathbf{F}^{2D} \in \mathbb{R}^d$, the Maximum-a-Posteriori~(MAP) estimator of 3D features $\mathbf{f}$ is
\begin{equation}
    \arg\max_\mathbf{f} p(\mathbf{f}|\mathbf{F}^{2D}) = \arg\max_\mathbf{f} p(\mathbf{F}^{2D}|\mathbf{f})\frac{p(\mathbf{f})}{p(\mathbf{F}^{2D})},
    \label{eq:map}
\end{equation}
which forms the basis of our method.

\paragraph{Preliminaries}

Given the LGS setting, we make the following key assumptions about the feature uplifting problem.

\begin{enumerate}[label=(\roman*)]
    \item \textbf{Uniform Prior}: The language features in CLIP space have uniform distribution, which is motivated by the efficient use of the CLIP feature space, due to training of a large amount of diverse data
    \begin{equation}
        p(\mathbf{f}) \propto \text{constant}.
        \label{eq:uniform_clip}
    \end{equation}
    
    \item \textbf{Conditional Independence of pixels}: Pixel observations are conditionally independent given the surface representation and 3D features $\mathbf{f}$
    \begin{equation}
        p(\mathbf{F}^{2D}|\mathbf{f}) = \prod_{v=1}^V \prod_{p \in P_v} p(\mathbf{f}_{p,v}^{2D}|\mathbf{f}).
        \label{eq:cond_ind}
    \end{equation}
    where $\mathbf{f}_{p,v}^{2D}$ denotes the 2D feature at pixel $p$ in view $v$.
    
    \item \textbf{Gaussian feature noise}: Viewpoint dependency of CLIP features can be modeled by a Gaussian distribution of feature noise
    \begin{equation}
        \epsilon_{p,v} \sim \mathcal{N}(\mathbf{0}, \sigma^2\mathbf{I}).
    \end{equation}
    
    \item \textbf{Alpha Blending Weights}: Alpha blending weights applied to each Gaussian $i$ at pixel $p$ in view $v$ can be interpreted as measures of occlusion between separate Gaussians in a given camera view
    \begin{equation}
        w_{i,p,v} = \alpha_i \prod_{j=1}^{i-1}(1-\alpha_j).
        \label{eq:alpha_blending}
    \end{equation}    
\end{enumerate}

\paragraph{Occam's LGS as a MAP estimator}
Our approach formulates feature uplifting as a MAP estimation problem, as stated in Equation~\ref{eq:map}. With the use of CLIP as language features, a model trained of large amounts of data is chosen. Under this condition, we make the assumption of a uniformly distributed feature space in Equation~\ref{eq:uniform_clip}. Under this condition, the LGS estimator is equivalent to the Maximum Likelihood~(ML)
\begin{equation}
    \arg\max_\mathbf{f} p(\mathbf{F}^{2D}|\mathbf{f}).
\end{equation}

This estimator can be evaluated separately for each 2D feature $\mathbf{f}_{p,v}^{2D}$ at pixel $p$ in view $v$ under conditional independence from Equation~\ref{eq:cond_ind}, given the surface representation and features $\mathbf{f}$ from the pretrained 3D Guassians
\begin{equation}
     \arg\max_\mathbf{f} p(\mathbf{F}^{2D}|\mathbf{f}) =  \arg\max_\mathbf{f} \prod_{v=1}^V \prod_{p \in P_v} p(\mathbf{f}_{p,v}^{2D}|\mathbf{f})
\end{equation}
where $P_v$ is the set of pixels in view $v$ on which Gaussian influences.

A major challenge in all LGS approaches is the viewpoint-dependent noise of 2D language features due to varying object appearance. Our approach formulates the ML estimation of LGS from noisy 2D features as IID normal distributed $\epsilon_{p,v} \sim \mathcal{N}(\mathbf{0}, \sigma^2\mathbf{I})$, representing the inherent uncertainty in the 2D features.

Hence, the 2D feature at pixel $p$ in view $v$, obtained from rendering the 3D Guassian features, is
\begin{equation}
    \mathbf{f}_{p,v}^{2D} = \sum_{k=1}^R w_{k,p,v}\mathbf{f}_k + \epsilon_{p,v},
\end{equation}
where $R$ is the number of Gaussians along the ray, $\epsilon_{p,v}$ is the language feature noise, and $w_{k,p,v}$ is the alpha blending weight for Gaussian $k$ at pixel $p$ in view $v$ defined in Equation~\ref{eq:alpha_blending}

To approach the ML estimation problem $\arg\max_\mathbf{f} p(\mathbf{F}^{2D}|\mathbf{f})$, we transform the problem to the log-likelihood domain
\begin{equation}
    \begin{aligned}
        &\log p(\mathbf{F}^{2D}|\mathbf{f}) = \log \prod_{v=1}^V \prod_{p \in P_v} p(\mathbf{f}_{p,v}^{2D}|\mathbf{f}) =\\
        &-\frac{1}{2\sigma^2} \sum_{v,p} \left\| \mathbf{f}_{p,v}^{2D} - \sum_{k=1}^R w_{k,p,v}\mathbf{f}_k \right\|^2 + \text{const},
    \end{aligned}
\end{equation}
and solve for the minimum by taking the derivative w.r.t. the 3D language feature $\mathbf{f}_i$ of Gaussian $i$
\begin{equation}
    \frac{\partial \log p}{\partial \mathbf{f}_i} = \frac{1}{\sigma^2} \sum_{v,p} w_{i,p,v} \left( \mathbf{f}_{p,v}^{2D} - \sum_{k=1}^R w_{k,p,v}\mathbf{f}_k \right) = 0.
\end{equation}
This leads to a formula where the weighted sum of observed 2D features (left side) relates to the interaction of Gaussians during the rendering process (right side)
\begin{equation}
    \Rightarrow \sum_{v,p} w_{i,p,v}\mathbf{f}_{p,v}^{2D} = \sum_{k=1}^R \sum_{v,p} w_{i,p,v}w_{k,p,v} \mathbf{f}_k.
\end{equation}

\paragraph{Practical Simplification}
Since our approach utilizes pretrained Gaussian Splats, we can make the assumption of sparse influence; i.e. for most views and pixels, one dominant Gaussian has significant influence (weight close to 1), which means that the rendering is dominated by non-overlapping Gaussians, leading to the following:
\begin{enumerate}
    \item[(a)] Negligible Cross-Talk: The product of weights from different Gaussians at the same pixel can be neglected compared to the dominant Gaussians, given the whole set of views
    $w_{i,p,v}w_{k,p,v} \approx 0$ for $i \neq k$. This eliminates the cross-Guassian terms in our equation.
    \item[(b)] Linearization: Since most weights are approximately binary (either close to 0 or close to 1), $w_{i,p,v}^2$ can be linearized as $\sum w_{i,p,v}^2 \approx \sum w_{i,p,v}$, given it is used inside a sum.
\end{enumerate}

With these properties of Gaussian Splats, the system of equations simplifies and can be written as the closed-form solution used in Ocamm's LGS
\begin{equation}
    \boxed{
        \mathbf{f}_i = \frac{\sum\limits_{v=1}^V \sum\limits_{p \in P_v} w_{i,p,v}\mathbf{f}_{p,v}^{2D}}{\sum\limits_{v=1}^V \sum\limits_{p \in P_v} w_{i,p,v}}.
    }
\end{equation}

This formula has an intuitive interpretation: the optimal 3D feature for each Gaussian is a weighted average of the 2D features observed at pixels where the Gaussian has influence, with weights proportional to the Gaussian's contribution to each pixel.



\section{Robustness to Feature Dimension Variation}

We also evaluate the localization accuracy across different feature dimensions on LERF dataset using the same feature compression method detailed in Sec~\ref{sec:featdim}. As shown by Table~\ref{tab:lerf-featdim-acc}, localization accuracy improves to 84.1 when feature dimension reduces to 32D, indicating that feature compression effectively reduces noisy signals.
\begin{table}[t]
    \centering
    \footnotesize
    \begin{tabular}{lccccc}
        \toprule
        \textbf{Dimensions}& 
        \hspace{-2pt}\textbf{Ramen}\hspace{-2pt} & \hspace{-2pt}\textbf{Figurines}\hspace{-2pt} & \hspace{-2pt}\textbf{Teatime}\hspace{-2pt} & \hspace{-2pt}\makecell{\textbf{Waldo}\\\textbf{Kitchen}}\hspace{-2pt} & \hspace{-2pt}\textbf{Overall}\hspace{-2pt} \\
        \midrule
        16-D & 69.0 & \textbf{94.9} & \textbf{90.9} & 78.6 & 83.4 \\
        32-D & 71.8 & 93.2 & \textbf{90.9} & \textbf{80.4} & \textbf{84.1} \\
        64-D & \textbf{73.2} & 93.2  & 86.4 & \textbf{80.4} & 83.3 \\
        128-D & 71.8 & 93.2  & 81.8 & 78.6 & 81.4 \\
        256-D & 74.7 & 93.2  & 86.4 & 78.6 & 83.2 \\
        512-D & 74.7 & 93.2  & 81.8 & \textbf{80.4} & 82.5 \\
        \bottomrule
    \end{tabular}
    \caption{Average accuracy with varying feature dimensions on LERF scenes}
    \label{tab:lerf-featdim-acc}
\end{table}

\section{Evaluation on 3DOVS Dataset}

\begin{figure}[!h]
    \centering
    \includegraphics[trim={6cm 2.5cm 10cm 3cm}, clip, width=0.5\textwidth]{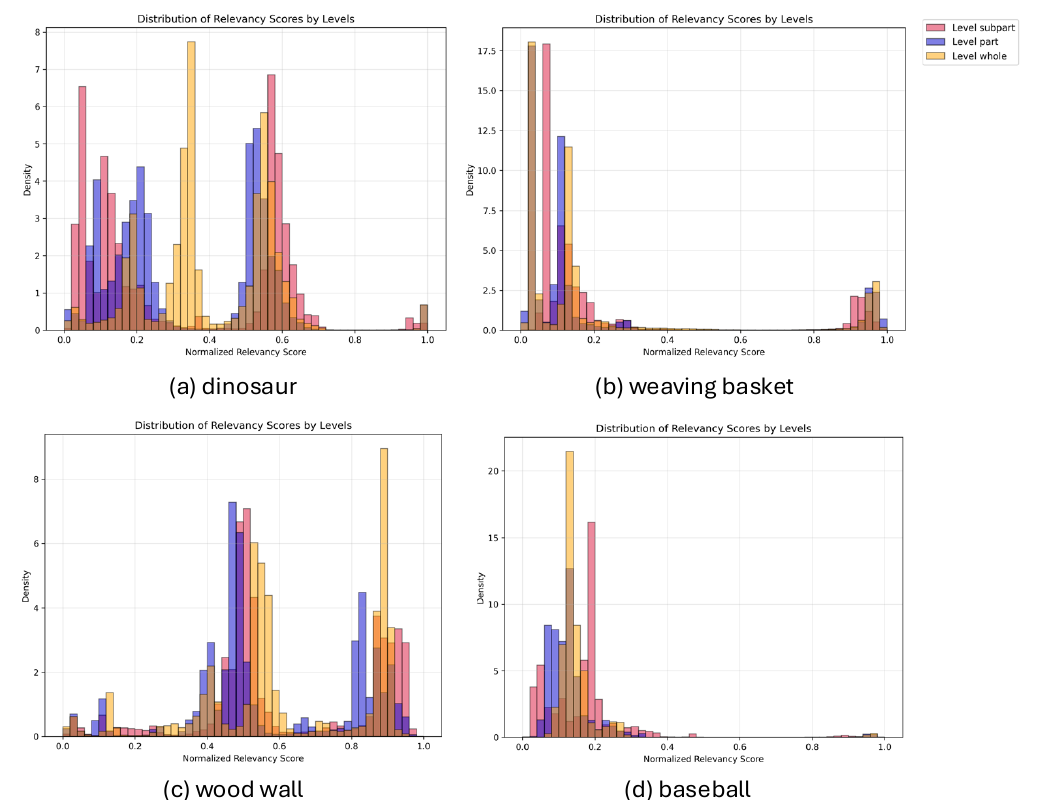}
    \caption{Distribution of relevancy scores across hierarchical levels (subpart, part, and whole) when applying different queries to room scene in the 3D-OVS dataset. Each color represents a distinct hierarchical level, illustrating the distribution of relevancy scores at different hierarchies. }
    \label{fig:distr}
\end{figure}
\begin{figure}[!h]
    \centering
    \vspace{-0.5cm}
    \includegraphics[trim={7cm 4cm 6cm 3cm}, clip, width=0.5\textwidth]{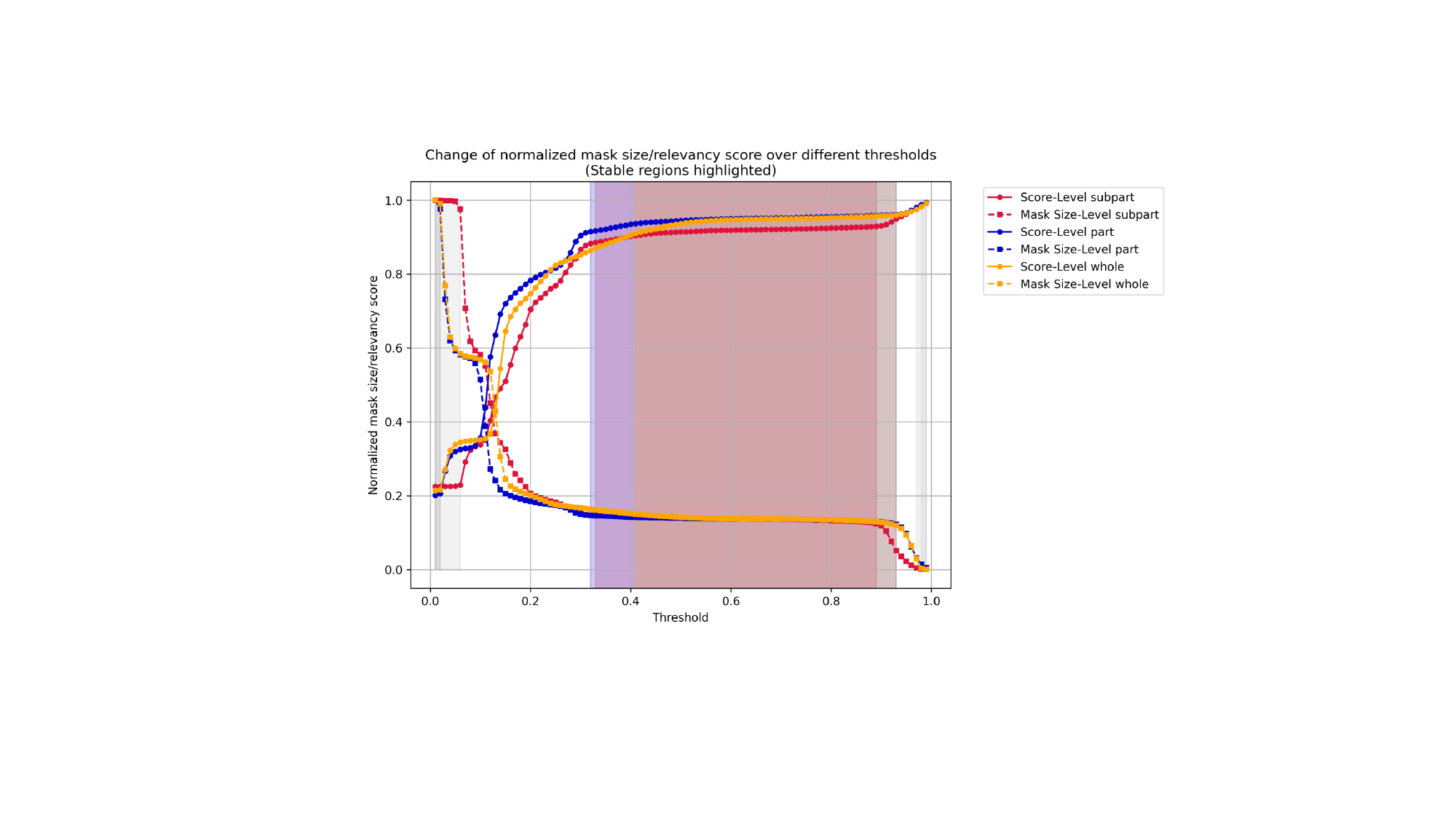}
    \caption{Relationship between score thresholds, normalized mask sizes, and relevancy scores for the "weaving basket" query in room scene. The plot shows how mask sizes decrease (downward trends with dotted lines) and average relevancy scores change (upward trends with smooth lines) as the threshold increases. Different hierarchical levels are represented by distinct colors, with the shaded region indicating the optimal stable region identified by our stability threshold.}
    \label{fig:grad}
\end{figure}

We observe that different object queries yield varying ranges of relevancy scores, which makes it challenging to set a consistent threshold for object segmentation across query types, as shown in Figure~\ref{fig:distr}. To address this limitation, we propose a dynamic thresholding approach. 
For each query, we evaluate a range of threshold values with a step size of 0.01. At each threshold, we create a binary mask by selecting regions where the relevancy score exceeds the current threshold value, and compute the average relevancy score within this masked area. These average relevancy scores at different thresholds are plotted in Figure~\ref{fig:grad}, shown by the increasing trend lines.

\begin{figure}[!h]
    \centering
    \includegraphics[trim={8cm 4cm 6cm 3cm}, clip, width=0.5\textwidth]{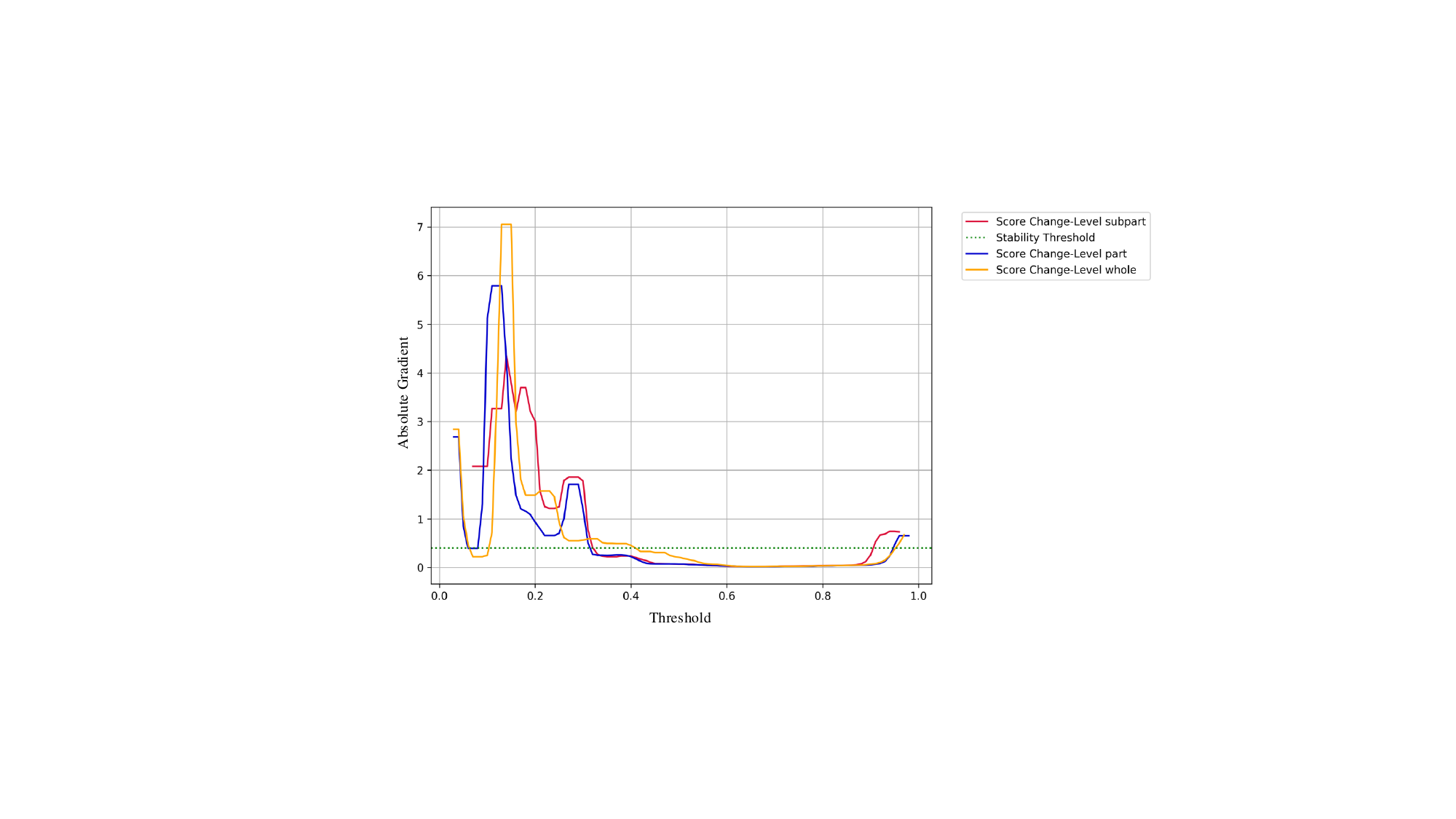}
    \caption{Visualization of the absolute gradient of average relevancy scores. The gradient measures the rate of change in average relevancy scores across different thresholds. The green dotted line indicates the stability threshold, with lower gradient values corresponding to more stable regions. This plot corresponds to the "weaving basket" query in the room scene.}
    \label{fig:grad_w_thres}
\end{figure}

Our goal is to identify a "stable region" where variations in the threshold minimally affect both the mask size and average relevancy score. Specifically, we find the region where the absolute gradient of the average relevancy score (the rate of change of average relevancy score with respect to different thresholds) is below the chosen stability threshold, as shown in Figure~\ref{fig:grad_w_thres}. We then select the middle value from this stable range as our final threshold for each different query, as illustrated in Figure~\ref{fig:grad}. The presence of a stable region indicates an object with consistent relevancy scores across threshold changes. In cases where multiple stable regions exist, we select the highest confidence prediction. We apply this improved evaluation methodology to the 3D-OVS dataset to ensure fair comparison across different query types.

\section{Additional Visualizations}
\subsection{LERF Visualization}
\begin{figure}[t]
    \centering
    \includegraphics[width=0.88\textwidth]{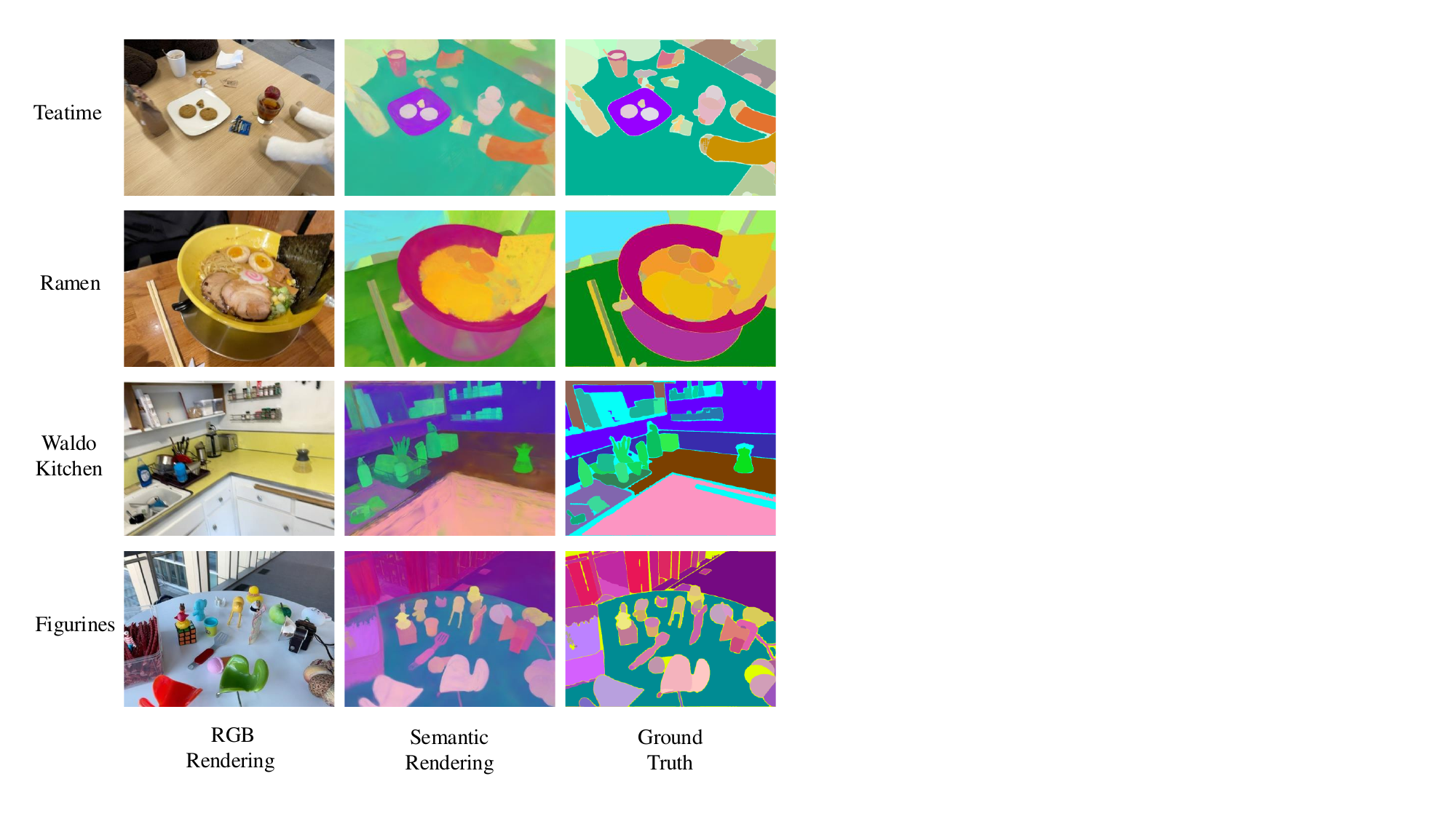}
    \caption{PCA visualization of semantic features on LERF Dataset. (a) Original RGB rendering. (b) PCA visualization of our rendered semantic feature maps. (c) Initial semantic feature maps extracted using CLIP from SAM segmentations. }
    \label{fig:lerf-vis3}
\end{figure}

We show additional visualization of the semantic features on LERF Dataset in Figure~\ref{fig:lerf-vis3}.

\subsection{3DOVS Visualization}
\begin{figure*}[t]
    \centering
    \includegraphics[trim={1cm 5cm 3cm 0.5cm}, clip, width=0.8\textwidth]{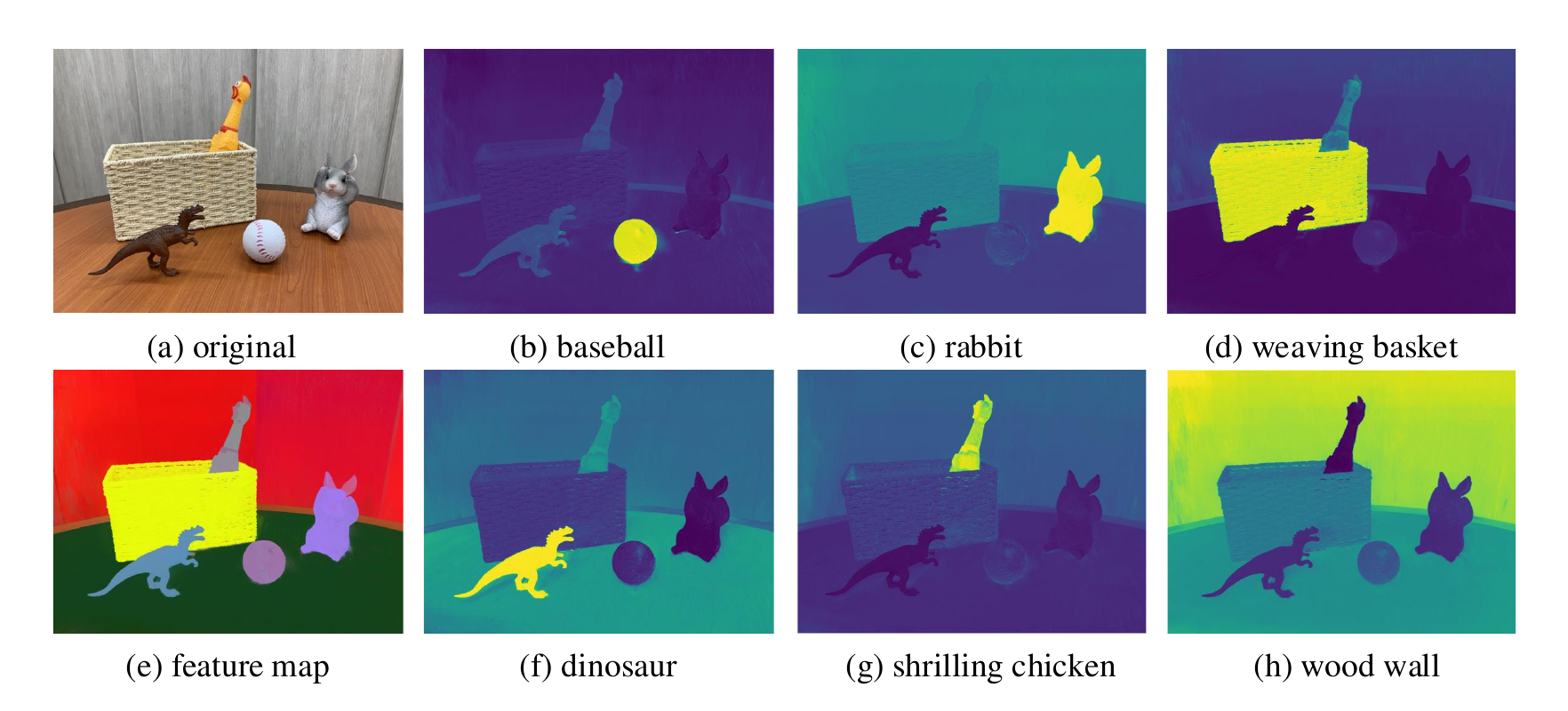}
    \caption{3D-OVS: room}
    \label{fig:room}
\end{figure*}

\begin{figure*}[t]
    \centering
    \includegraphics[trim={1cm 5cm 3cm 0.5cm}, clip, width=0.8\textwidth]{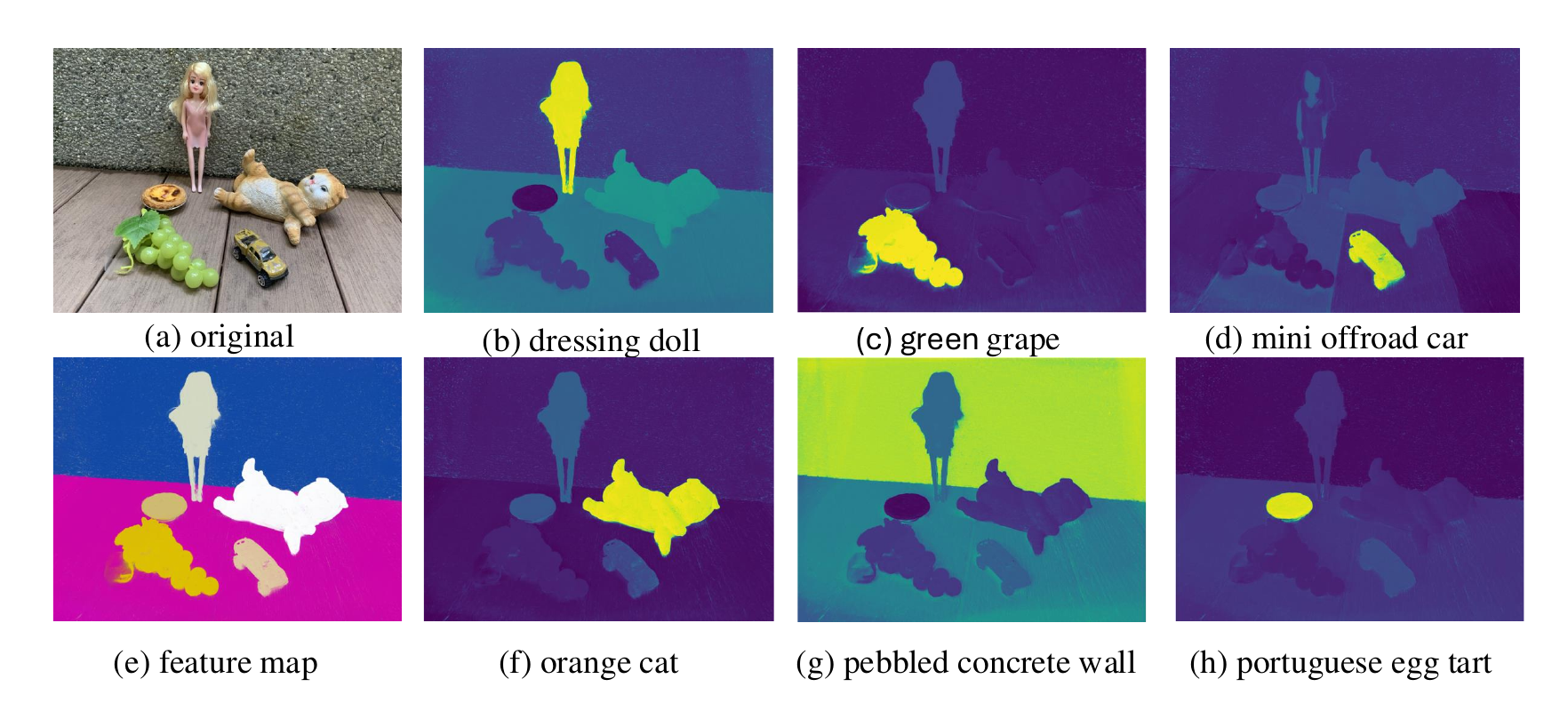}
    \caption{3D-OVS: bench}
    \label{fig:bench}
\end{figure*}

\begin{figure*}[t]
    \centering
    \includegraphics[trim={1cm 5cm 3cm 0.5cm}, clip, width=0.8\textwidth]{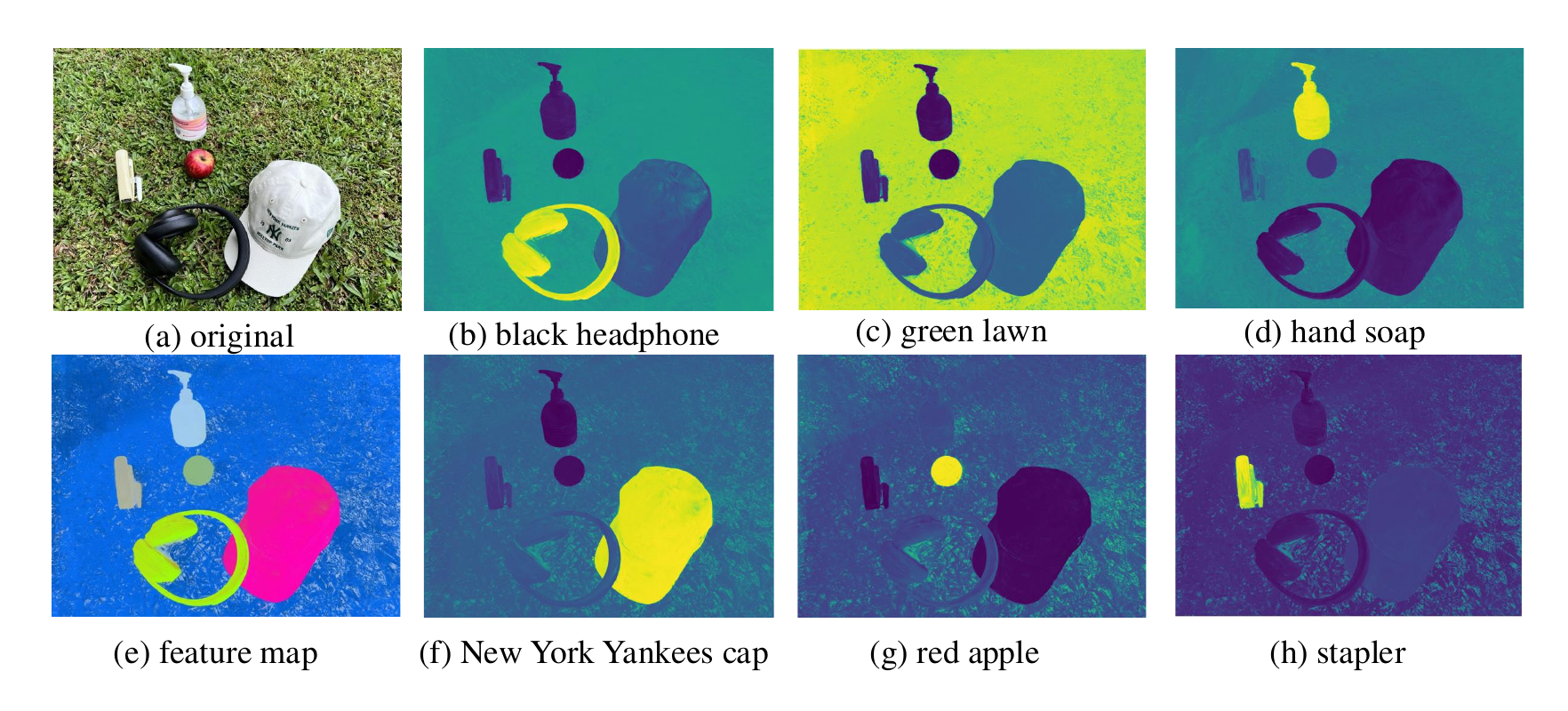}
    \caption{3D-OVS: lawn}
    \label{fig:lawn}
\end{figure*}

We show more examples on the 3D-OVS dataset in Figure~\ref{fig:room},~\ref{fig:bench}, and ~\ref{fig:lawn}.

\end{document}